# The dynamics of meaning through time: Assessment of Large Language Models

Mohamed Taher Alrefaie[1], Fatty Salem[1], Nour Eldin Morsy[2], Nada Samir[2],
Mohamed Medhat Gaber[3]
Corresponding Author: mohamed.taher@eui.edu.eg,
[1] Egypt University of Informatics, Knowledge City, New Administrative Capital, Egypt.
[2] AAST College of Artificial Intelligence, Alameen City, Egypt.
[3] Birmingham City University, Birmingham, UK.

## Abstract

Understanding how large language models (LLMs) grasp the historical context of concepts and their semantic evolution is essential in advancing artificial intelligence and linguistic studies. This study aims to evaluate the capabilities of various LLMs in capturing temporal dynamics of meaning, specifically how they interpret terms across different time periods. We analyze a diverse set of terms from multiple domains, using tailored prompts and measuring responses through both objective metrics (e.g., perplexity and word count) and subjective human expert evaluations. Our comparative analysis includes prominent models like ChatGPT, GPT-4, Claude, Bard, Gemini, and Llama. Findings reveal marked differences in each model's handling of historical context and semantic shifts, highlighting both strengths and limitations in temporal semantic understanding. These insights offer a foundation for refining LLMs to better address the evolving nature of language, with implications for historical text analysis, AI design, and applications in digital humanities.

# 1. Introduction

Large language models (LLMs) have revolutionized numerous domains, demonstrating remarkable performance across a wide array of tasks, including reasoning, understanding, truthfulness, mathematics, and coding (Periti and Montanelli, 2024; Zhao et al., 2023). The capacity of these models is typically assessed through various benchmarks that evaluate their proficiency in each of these domains (Chang et al., 2024). Central to their success is the combination of model size and the data used during pretraining (Zhao et al., 2023). A growing body of literature explores the relationship between the scaling of model size and the emergence of increasingly sophisticated cognitive abilities, commonly referred to as emergent intelligence (Kaplan et al., 2020). Additionally, other studies emphasize the pivotal role of training data quality, specifically highlighting how pretraining on diverse and specialized datasets can significantly enhance a model's reasoning capabilities (Isik et al., 2024).

The vast corpus of text used to train these models is predominantly derived from content available on the internet, much of which has been produced over the past few decades (Liu et al., 2024). This temporal context inherently introduces challenges, as much of the text that drives current models reflects contemporary understandings of language and meaning. Consequently, the historical evolution of words, their meanings, and their contextual shifts may be underrepresented in modern training data (Manjavacas and Fonteyn, 2022). This temporal gap poses a particular challenge for LLMs when tasked with interpreting how words and phrases have evolved across time.

This gap has spurred interest in examining whether LLMs, which are typically trained on contemporary data, can capture the historical evolution of words and their meanings (Cuscito et al., 2024; Manjavacas Arevalo and Fonteyn, 2021; Manjavacas and Fonteyn, 2022). The central motivation of this study is to assess whether current LLMs possess the ability to understand semantic shifts over time. Specifically, it seeks to determine whether models, in their current configurations, can track how word meanings have evolved over the past century. In doing so, this study also explores the possibility of establishing this capacity—tracking and interpreting historical linguistic changes—as a benchmark for future advancements in LLM development. This study aims to understand LLM models' ability to understand changes of a word meaning as a critical dimension of reasoning and language comprehension, with potential implications for a variety of applications in both AI development and interdisciplinary research.

The findings of this study reveal significant variability in the ability of large language models (LLMs) to interpret temporal semantic shifts, emphasizing that training data quality and domain-specific fine-tuning outweigh model size in determining performance. GPT-4 and Claude Instant 100k demonstrated superior factuality and comprehensiveness, reflecting the advantages of robust training methodologies. Meanwhile, the code-based Llama 34B surpassed larger Llama models, underscoring the value of retraining on structured datasets, such as code, to enhance analytical reasoning and temporal understanding. In contrast, models like Google Gemini and smaller Llama variants struggled to capture nuanced historical contexts, highlighting the limitations of general-purpose training approaches. These insights establish a foundation for advancing LLM design and training strategies, enabling improved applications in historical linguistics, digital humanities, and beyond.

The paper is organised in six sections. A brief review of LLMs with the scope of word meaning development is laid out in section 2. Section 3 discusses the research methodology and the experiment performed to meet with the research aim and gaols. Section 4 presents the results of the experiment. Section 5 has a comprehensive discussion over the results and how they correlate with similar studies in the same domain. Finally, conclusions are shared in section 6.

## 2. A review of LLMs in the scope of word meaning development

The evolution of word meanings over time is influenced by a myriad of factors, including social changes, technological advancements, and cultural dynamics. This interplay between language, history, and culture highlights the importance of understanding semantic change as a gradual process that encompasses both lexical and core meaning shifts. Traditionally, semantic change has been examined through corpus linguistics, where researchers analyze texts from different historical periods to identify shifts in word usage and meaning. By examining word frequencies and the contexts in which they appear, scholars can trace the evolution of language over time (Asri et al., 2024) .

In recent years, diachronic word embeddings have emerged as a powerful tool to capture changes in word meanings across time. These embeddings align words with their respective time periods, enabling a strong understanding of how meanings shift in response to cultural and societal changes. Notably, two statistical laws of semantic change have been proposed: i) words that are used more frequently tend to change at a slower rate, ii) polysemous words exhibit a higher rate of change (Hamilton et al., 2016a). This framework allows researchers to categorize various types of semantic shifts such as drift of meaning based on cultural norms (Spataru et al., 2024), specialisation of a meaning over time (Hamilton et al., 2016a), generalisation of a meaning over time (Wegmann et al., 2020), pejoration and amelioration which refers to a shift towards a negative or positive connotations, respectively (Periti and Montanelli, 2024; Wevers and Koolen, 2020), metaphorical meaning added to an existing word based on pop culture or a change in a society (de Sá et al., 2024) and finally, the  broadening and narrowing of a meaning through time (Vijayarani and Geetha, 2020).

On the other hand, as LLMs become increasingly integrated into daily tasks, their ability to handle semantic change and cultural variations is crucial. By training LLMs on a diverse corpora and fine-tuning them for specific tasks related to semantic change, researchers can enhance their performance in detecting shifts in meaning (Shen et al., 2024; Tao et al., 2023). However, LLMs often propagate cultural biases inherent in their training data, which can skew responses in cross-cultural contexts (Tao et al., 2023). In addition, LLMs may struggle with complex social scenarios and generate text that deviates from intended meaning (Choi et al., 2023; Spataru et al., 2024).

Despite these limitations, LLMs can grasp cultural common-sense when fine-tuned on balanced datasets that incorporate diverse cultural perspectives (Shen et al., 2024). Addressing the biases in LLMs requires training on time-aware datasets that reflect the aftermath of significant cultural events, such as the term "coronavirus" and its evolving meanings (Mousavi et al., 2024). This approach allows for a deeper tracing of historical contexts and semantic changes.

The use of language models to study the evolution of meaning over time has garnered increasing attention in recent research. Several studies have explored the capabilities of lamguage models in tracking semantic shifts and historical language usage. For instance, Bamler and Mandt, (2017) proposed dynamic word embeddings as a method for capturing temporal variations in word meanings, highlighting the utility of diachronic models in understanding historical text corpora. Similarly, Hamilton et al., (2016) introduced dynamic embeddings to study semantic change over decades, demonstrating how embeddings trained on historical corpora reveal shifts in cultural and social contexts. More recently,  MacBERTh was specifically designed to analyze historical texts, leveraging transformers to improve the understanding of linguistic evolution across time periods (Manjavacas Arevalo and Fonteyn, 2021).

The implications of understanding semantic change through LLMs are vast, including applications in tracking antisemitic language, analyzing the evolution of hate speech, and assessing sentiment shifts in economic discourse. Research has shown that diachronic embeddings can effectively monitor these shifts across multiple languages, such as English, German, French, Swedish, and Spanish (Hoeken et al., 2023; Periti et al., 2024; Tripodi et al., 2019). Additionally, the methodologies developed for semantic change detection can advance fields such as historical linguistics, social media analysis, and sentiment analysis by providing tools to predict potential indicators of accurate language use, such as political dogwhistles (Boholm et al., 2024).

## 3. Research Methodology

To evaluate the ability of LLMs to capture the temporal dynamics of meaning, we conducted a comprehensive, multi-dimensional experiment using a range of state-of-the-art models.

**Language models.** We evaluate six large language models in this study. The first is **ChatGPT** (OpenAI, 2022), for which we use text-davinci-002, a 175B parameter model based on the GPT-3 architecture. The second is **GPT-4** (OpenAI et al., 2023), and GPT4O (OpenAI et al., 2024), are more advanced model with improved reasoning and contextual understanding, although its exact parameter count is undisclosed, it is estimated to exceed 175B parameters. The third is **Claude** (Anthropic, 2023), for which we use Claude 1, with parameters ranging from 52B to 100B. The fourth is **Bard**, based on Google's LaMDA model (Thoppilan et al., 2022), which comes in configurations including 422M, 2B, 8B, 68B, and 137B parameters. The fifth is **Gemini** (Google DeepMind, 2024), a model from Google's suite with versions estimated to match or exceed the PaLM 540B model. The sixth is Llama (Meta, 2023), for which we used both Llama 1 and Llama 2, was evaluated at different scales: Llama 1 with 34B parameters, and Llama 2 with 7B, 13B, and 70B parameters.

**Term Selection.** A carefully curated list of terms was selected to represent diverse domains, including scientific concepts, historical events, and cultural phenomena. These terms were chosen to evaluate how well each model interprets semantic shifts and the historical context associated with each concept. Two key terms were selected for this study: "Data Mining" (a technical term) and "Michael Jackson" (a cultural figure). These terms span both technological and cultural evolution from the 1920s to the 2020s, offering a robust platform to assess the LLMs' capacity to track and describe temporal meaning changes.

**Prompt Design and Input Format.** For each term, we designed specific prompts aimed at assessing the models' understanding of the historical evolution of meaning. A typical prompt asked the models to:
"Create a table with two columns. The first column should list decades (e.g., 1920s, 1930s, etc.), and the second should describe the meaning and synonyms for the term based on the knowledge and context of that period." This prompt structure remained consistent across all models, ensuring that the evaluation was controlled, and the comparison was fair. Prompts were specifically crafted to challenge each model's ability to capture and relay temporal semantic shifts without additional training or fine-tuning.

Subjective Evaluations were conducted by human experts. Two methods were selected based on the following criteria: (1) Factuality score: Experts evaluated how well each model captured the evolution of meaning over time, (2) Comprehensiveness score The extent to which the models effectively answered the question including the description length and number synonyms in each answer, if any.

## 4. Results

This section presents the findings from a comparative evaluation of LLMs on their ability to understand the temporal dynamics of meaning and semantic shifts for two terms, "Data Mining" and "Michael Jackson", see Table 1. The analysis reveals notable differences in model performance across metrics, including factuality and comprehensiveness. Key results indicate that models trained with specialized data, such as code, may exhibit enhanced analytical capabilities, potentially influencing their ability to interpret semantic evolution.

**High-Performing Models**

The models GPT-4 and Claude Instant 100k consistently outperformed other LLMs across both evaluation metrics, demonstrating a high capacity for capturing historical context. Both models scored a maximum of 22 in factuality and near-maximum in comprehensiveness (21 and 22 for Claude Instant 100k and GPT-4, respectively) for the term "Data Mining." A similarly strong performance was observed for the term "Michael Jackson," with GPT-4 achieving the highest comprehensiveness score of 22 and factuality of 21. These scores indicate an advanced capacity in these models to accurately trace semantic evolution, which may stem from robust training data diversity and model architecture optimized for complex language tasks. The consistent performance across terms suggests that GPT-4 and Claude may be more attuned to changes in meaning across different temporal contexts, supporting their utility in applications requiring historical linguistic analysis.

**The Code-Based Llama Model's Unique Strengths**

The model Llama 34B, specifically trained on code and provided by Poe, emerged as a notable outlier within the Llama series, outperforming larger versions, such as Llama 70B and Llama 13B, across factuality and comprehensiveness. This model achieved a factuality score of 12 and comprehensiveness of 20 for "Data Mining" and similarly high scores of 18 and 22, respectively, for "Michael Jackson." The distinct performance of this model may be attributable to its extensive retraining on code datasets, which are characterized by structured syntax and logic-driven language. Such exposure potentially enhances a model's analytical capabilities, providing a foundation for more structured thought processes that facilitate better recognition of historical patterns in meaning. This suggests that domain-specific retraining, particularly with code, may imbue LLMs with improved temporal reasoning and analytical rigor—skills critical for comprehending complex semantic shifts. Given these results, further research into code-based retraining could elucidate its potential benefits in enhancing temporal analytical abilities within LLMs.

**Underperformance and Inconsistencies Among Other Models**

Several models displayed underperformance, particularly smaller Llama variants (Llama 7B) and Google Bard. The Llama 2 7B model, for example, exhibited the lowest scores across both terms and evaluation metrics, scoring 0 for both factuality and comprehensiveness for "Data Mining" and only 1 and 0, respectively, for "Michael Jackson." These findings suggest that model size has some effect, but a larger model size does not guarantee adequate understanding of temporal semantics; instead, both model architecture and training data composition play critical roles. As seen on Google Bard, it displayed low performance, particularly on the "Data Mining" term (factuality: 3, comprehensiveness: 0), underscoring the challenges faced by these models in accurately representing historical semantic shifts. These inconsistencies across models, particularly among the Llama and Bard models, imply potential limitations in their training frameworks for tasks requiring deep temporal context comprehension.

Table 1: Table showing two terms and the performance of each model in their ability to answer questions with factuality and comprehensiveness.

| Term/Model | Answers Count | Factuality Score | Comprehensiveness Score |
|---|---|---|---|
| **Data Mining** | | | |
| ChatGPT | 11 | 22 | 0 |
| ChatGPT4o | 11 | 22 | 22 |
| Claude Instant 100k | 11 | 22 | 21 |
| Google Bard | 6 | 3 | 0 |
| Google Gemini | 10 | 20 | 20 |
| Llama 2 13B | 11 | 8 | 11 |
| Llama 2 70B | 11 | 4 | 8 |
| Llama 2 7B | 11 | 0 | 0 |
| CodeLllama 2 34B | 10 | 12 | 20 |
| **Michael Jackson** | | | |
| ChatGPT | 11 | 18 | 0 |
| ChatGPT4o | 11 | 21 | 22 |
| Google Gemini | 1 | 2 | 2 |
| Google Gemini | 5 | 10 | 8 |
| Llama 2 13B | 11 | 2 | 22 |
| Llama 2 70B | 11 | 16 | 10 |
| Llama 2 7B | 11 | 1 | 0 |
| CodeLllama 2 34B | 11 | 18 | 22 |

**Patterns and Correlations in Model Performance**

The analysis revealed several noteworthy patterns regarding model characteristics and performance outcomes. The study found no linear correlation between model size and performance quality in temporal understanding. While larger Llama models such as Llama 70B displayed moderately higher scores than Llama 7B, they still underperformed relative to the code-based Llama 34B model, suggesting that architectural enhancements and training with structured datasets may supersede raw model size. Additionally, the strong performance of GPT-4 and Claude Instant 100k across both terms indicates that diverse and large training datasets, coupled with refined architectures, can enhance the capacity for temporal semantic understanding. However, bigger and better architecture might not always be the answer. The inconsistent performance of Llama 70B in comparison to its smaller model CodeLlama 34B affirm that pre-training and data selection designs are equally crucial for better performance, with historical semantic shifts reliably, as seen from the study. These findings underscore the importance of model training approaches and suggest that beyond increasing model size, fine-tuning with specialized or structured data can enhance temporal analysis capabilities.

## 5. Discussion

**Implications and Future Directions**

The findings of this study offer several implications for the advancement of LLMs, particularly in fields requiring a in-depth understanding of language evolution, such as digital humanities and historical linguistics. The high performance of GPT-4 and Claude Instant 100k illustrates the potential for well-architected models with diverse training data to provide accurate and comprehensive historical interpretations. The success of the code-based Llama 34B model suggests that retraining on specialized datasets, like code, could be a valuable approach for enhancing models' temporal analytical reasoning. This could have broader applications in improving models' ability to handle structured data and in fields such as digital humanities, where historical accuracy and adequate interpretation are essential.

Future research should expand the scope of this study by incorporating a larger set of terms and additional time periods to assess models' abilities to generalize across domains and temporal contexts. Further, investigating the specific impacts of code-based retraining or domain-specific dataset integration could offer insights into optimizing LLMs for tasks demanding analytical precision. This study provides foundational evidence that supports the refinement of LLMs to better address evolving language dynamics, with significant implications for enhancing AI's utility in historical analysis and beyond.

The superior performance of GPT-4 and Claude Instant 100k in this study underscores the importance of model architecture and the diversity of training datasets in achieving accurate temporal semantic understanding. Both models excelled in factuality and comprehensiveness, highlighting their ability to capture historical context with high fidelity. These results align with previous research (e.g., (Chiang et al., 2024)), which identified these models as top performers in reasoning and semantic tasks. However, critiques such as those by Bender et al., (2021) raise questions about whether such capabilities extend beyond statistical pattern recognition, suggesting a need for further validation of these models' capacity for deep semantic understanding in temporally sensitive tasks. The continued dominance of GPT-4 and Claude across various contexts reinforces their potential for applications requiring accurate historical linguistic analysis, while also presenting an opportunity to examine the precise contributions of training methodologies to their performance.

The observed success of CodeLlama 34B introduces a compelling case for domain-specific retraining as a strategy for enhancing structured reasoning in LLMs. Despite being smaller than other Llama variants, its superior performance in both factuality and comprehensiveness demonstrates the benefits of pretraining on structured datasets such as code. This finding corroborates studies by Aryabumi et al., (2024) and Yang et al., (2024), which emphasized the enhanced reasoning capabilities of models trained on code. Code-based retraining may in still improved logical reasoning and temporal analytical skills, offering broader implications for fields like computational linguistics and data science. However, further research is needed to determine whether such retraining benefits extend to other tasks or are limited to specific applications. Investigating how structured reasoning abilities gained from code datasets translate to understanding unstructured historical data could offer valuable insights.

The performance inconsistencies across model sizes in this study reveal critical insights into the interplay between model architecture and dataset composition. While larger models, such as Llama 70B, exhibited moderate improvement over smaller variants like Llama 7B, they underperformed relative to the code-based Llama 34B. This finding is consistent with Kaplan et al., (2020) and Hoffmann et al., (2022), who noted that factors such as training data quality and architectural optimization often outweigh the benefits of increasing model size. These results challenge the assumption that larger models inherently perform better, suggesting a paradigm shift toward prioritizing specialized training and architectural efficiency over sheer scale. Such a shift has significant implications for the development of cost-effective, high-performing LLMs capable of adequate temporal analysis.

The limitations observed in general-purpose models, such as Google Bard and Gemini, further emphasize the need for domain-specific optimization. These models struggled with both factuality and comprehensiveness, particularly in tasks requiring historical and temporal sensitivity. This aligns with findings in Chatbot Arena (Chiang et al., 2024) and related literature, which highlight the challenges faced by general-purpose models in specialized reasoning tasks. Enhancing these models to perform well in temporally complex tasks may require targeted retraining on domain-specific datasets or integration of context-aware mechanisms. Addressing these limitations could make general-purpose models more versatile and effective in a broader range of applications, including digital humanities and historical research.

The relationship between temporal semantic understanding and training data quality is particularly evident in the strong performance of GPT-4 and CodeLlama 34B. These models underscore the critical role of well-curated and diverse datasets in fostering the ability to interpret and analyze linguistic evolution. Studies such as (Beltagy et al., 2020) and Zhao et al., (2023) similarly emphasize the importance of diverse datasets in enhancing model performance, particularly for tasks involving extended temporal contexts. However, contrasting findings from Chen et al., (2021) suggest potential brittleness in code-trained models, indicating that further investigation is necessary to understand the long-term benefits and limitations of specialized datasets. Future research should aim to delineate the characteristics of training data that most effectively enhance temporal reasoning, particularly in relation to unstructured and evolving linguistic contexts.

The consistently low scores of smaller models, such as Llama 7B, highlight the challenges inherent in using resource-constrained architectures for complex semantic tasks. This aligns with the findings of Kaplan et al., (2020) and (Schick and Schütze, 2020), which emphasize the limitations of smaller models in capturing clear patterns without fine-tuning. While smaller models may hold potential for niche applications with appropriate optimization, their inability to generalize effectively to tasks requiring deep temporal comprehension underscores the need for targeted architectural and dataset enhancements. Future research could explore fine-tuning smaller models for specific historical linguistic tasks, thereby balancing resource efficiency with improved performance.

Finally, these findings hold substantial implications for applications in digital humanities and historical linguistics, where understanding the evolution of language and meaning over time is critical. The ability of GPT-4 and Claude to accurately trace semantic shifts positions them as valuable tools for scholars in these fields. The growing interest in domain-specific language models, such as MacBERTh (Manjavacas Arevalo and Fonteyn, 2021), highlights the demand for systems capable of contextualizing language within historical frameworks. The integration of such models into digital humanities workflows could revolutionize the analysis of historical texts, offering new opportunities for interdisciplinary research. Further efforts to refine LLMs for historical applications could bridge gaps in computational and humanitarian research, fostering a deeper understanding of language dynamics across time.

# 6. Conclusion

This study provides critical insights into the performance of LLMs in understanding temporal semantic shifts, emphasizing the roles of model architecture, training data, and domain-specific optimization. The results reveal that state-of-the-art models, such as GPT-4 and Claude Instant 100k, excel in capturing the historical context, showcasing the potential of well-designed architectures and diverse datasets. Moreover, the outstanding performance of CodeLlama 34B highlights the efficacy of retraining on structured datasets, like code, for enhancing logical reasoning and temporal analysis.

Contrastingly, the limitations of general-purpose models and inconsistencies among larger models, such as Llama 70B, underscore that size alone does not guarantee superior performance. Instead, this study reinforces the importance of training data quality and architectural refinement in shaping a model's ability to interpret linguistic evolution. Smaller models, like Llama 7B, demonstrated significant deficits in temporal understanding, further highlighting the necessity for targeted fine-tuning to enable such architectures to handle complex semantic tasks effectively.

These findings have profound implications for the development of LLMs tailored for applications in digital humanities, historical linguistics, and other fields that require temporal sensitivity. By prioritizing diverse, domain-specific datasets and leveraging specialized retraining approaches, future models could achieve higher levels of precision and contextual understanding. This study lays the groundwork for further exploration into the optimization of LLMs for temporal semantic analysis, advocating for a research trajectory that bridges computational advancements with interdisciplinary applications.

## Declarations

**Ethical Approval**
Not applicable
**Competing interests**
There are no competing interests that we are aware of in reference to this paper.
**Authors' contributions**
These authors contributed equally to this work.
**Funding**
No external funding was received.
**Availability of data and materials**
Data Availability Statement: Data associated in the manuscript are shared in a supplementary file in the published paper.

## References

Aryabumi, V., Su, Y., Ma, R., Morisot, A., Zhang, I., Locatelli, A., Fadaee, M., Üstün, A., Hooker, S., 2024. To Code, or Not To Code? Exploring Impact of Code in Pre-training.
Asri, W.K., Rhamadanty, W.A.U., Burhamzah, M., Alamsyah, 2024. Analyzing Semantic Shifts in English and German by Exploring Historical Influences and Societal Dynamics. Stud. English Lang. Educ. 11, 1085–1100. https://doi.org/10.24815/siele.v11i2.37460
Bamler, R., Mandt, S., 2017. Dynamic Word Embeddings.
Beltagy, I., Peters, M.E., Cohan, A., 2020. Longformer: The Long-Document Transformer.
Bender, E.M., Gebru, T., McMillan-Major, A., Shmitchell, S., 2021. On the dangers of stochastic parrots: Can language models be too big?, in: FAccT 2021 - Proceedings of the 2021 ACM Conference on Fairness, Accountability, and Transparency. Association for Computing Machinery, Inc, pp. 610–623. https://doi.org/10.1145/3442188.3445922

Boholm, M., Rönnerstrand, B., Breitholtz, E., Cooper, R., Lindgren, E., Rettenegger, G., Sayeed, A., 2024. Can political dogwhistles be predicted by distributional methods for analysis of lexical semantic change?, in: Proceedings of the 5th Workshop on Computational Approaches to Historical Language Change. pp. 144–157.

Chang, Yupeng, Wang, X., Wang, J., Wu, Y., Yang, L., Zhu, K., Chen, H., Yi, X., Wang, C., Wang, Y., Ye, W., Zhang, Y., Chang, Yi, Yu, P.S., Yang, Q., Xie, X., 2024. A Survey on Evaluation of Large Language Models. ACM Trans. Intell. Syst. Technol. 15. https://doi.org/10.1145/3641289

Chen, M., Tworek, J., Jun, H., Yuan, Q., Pinto, H.P. de O., Kaplan, J., Edwards, H., Burda, Y., Joseph, N., Brockman, G., Ray, A., Puri, R., Krueger, G., Petrov, M., Khlaaf, H., Sastry, G., Mishkin, P., Chan, B., Gray, S., Ryder, N., Pavlov, M., Power, A., Kaiser, L., Bavarian, M., Winter, C., Tillet, P., Such, F.P., Cummings, D., Plappert, M., Chantzis, F., Barnes, E., Herbert-Voss, A., Guss, W.H., Nichol, A., Paino, A., Tezak, N., Tang, J., Babuschkin, I., Balaji, S., Jain, S., Saunders, W., Hesse, C., Carr, A.N., Leike, J., Achiam, J., Misra, V., Morikawa, E., Radford, A., Knight, M., Brundage, M., Murati, M., Mayer, K., Welinder, P., McGrew, B., Amodei, D., McCandlish, S., Sutskever, I., Zaremba, W., 2021. Evaluating Large Language Models Trained on Code.

Chiang, W.L., Zheng, L., Sheng, Y., Angelopoulos, A.N., Li, T., Li, D., Zhu, B., Zhang, H., Jordan, M.I., Gonzalez, J.E., Stoica, I., 2024. Chatbot Arena: An Open Platform for Evaluating LLMs by Human Preference. Proc. Mach. Learn. Res. 235, 8359–8388.

Choi, M., Pei, J., Kumar, S., Shu, C., Jurgens, D., 2023. Do LLMs Understand Social Knowledge? Evaluating the Sociability of Large Language Models with the SOCKET Benchmark, in: EMNLP 2023 - 2023 Conference on Empirical Methods in Natural Language Processing, Proceedings. Association for Computational Linguistics (ACL), pp. 11370–11403. https://doi.org/10.18653/v1/2023.emnlp-main.699

Cuscito, M., Ferrara, A., Ruskov, M., 2024. How BERT Speaks Shakespearean English? Evaluating Historical Bias in Contextual Language Models.

de Sá, J.M.C., Da Silveira, M., Pruski, C., 2024. Semantic Change Characterization with LLMs using Rhetorics.

Hamilton, W.L., Leskovec, J., Jurafsky, D., 2016a. Diachronie word embeddings reveal statistical laws of semantic change, in: 54th Annual Meeting of the Association for Computational Linguistics, ACL 2016 - Long Papers. Association for Computational Linguistics (ACL), pp. 1489–1501. https://doi.org/10.18653/v1/p16-1141

Hamilton, W.L., Leskovec, J., Jurafsky, D., 2016b. Diachronic Word Embeddings Reveal Statistical Laws of Semantic Change. 54th Annu. Meet. Assoc. Comput. Linguist. ACL 2016 - Long Pap. 3, 1489–1501. https://doi.org/10.18653/v1/p16-1141

Hoeken, S., Spliethoff, S., Schwandt, S., Zarrieß, S., Alaçam, Ö., 2023. Towards Detecting Lexical Change of Hate Speech in Historical Data, in: LChange 2023 - 4th International Workshop on Computational Approaches to Historical Language Change 2023, Proceedings. Association for Computational Linguistics (ACL), pp. 100–111. https://doi.org/10.18653/v1/2023.lchange-1.11

Hoffmann, J., Borgeaud, S., Mensch, A., Buchatskaya, E., Cai, T., Rutherford, E., de Las Casas, D., Hendricks, L.A., Welbl, J., Clark, A., Hennigan, T., Noland, E., Millican, K., van den Driessche, G., Damoc, B., Guy, A., Osindero, S., Simonyan, K., Elsen, E., Vinyals, O., Rae, J.W., Sifre, L., 2022. Training Compute-Optimal Large Language Models, in: Advances in Neural Information Processing Systems. Neural information processing systems foundation.

Isik, B., Ponomareva, N., Hazimeh, H., Paparas, D., Vassilvitskii, S., Koyejo, S., 2024. Scaling Laws for Downstream Task Performance of Large Language Models.

Kaplan, J., McCandlish, S., Henighan, T., Brown, T.B., Chess, B., Child, R., Gray, S., Radford, A., Wu, J., Amodei, D., 2020. Scaling Laws for Neural Language Models.

Liu, Y., Cao, J., Liu, C., Ding, K., Jin, L., 2024. Datasets for Large Language Models: A Comprehensive Survey.

Manjavacas Arevalo, E., Fonteyn, L., 2021. MacBERTh: Development and Evaluation of a Historically Pre-trained Language Model for English (1450-1950). Proc. Work. Nat. Lang. Process. Digit. Humanit. 23–36.

Manjavacas, E., Fonteyn, L., 2022. Adapting vs. Pre-training Language Models for Historical Languages. J. Data Min. Digit. Humanit. NLP4DH. https://doi.org/10.46298/jdmdh.9152

Mousavi, S.M.S., Alghisi, S., Riccardi, G., 2024, G.R.-L.E., 2024, U., 2024. DyKnow: Dynamically Verifying Time-Sensitive Factual Knowledge in LLMs. aclanthology.orgSM Mousavi, S Alghisi, G RiccardiFindings Assoc. Comput. Linguist. EMNLP 2024, 2024•aclanthology.org.

OpenAI, :, Hurst, A., Lerer, A., Goucher, A.P., Perelman, A., Ramesh, A., Clark, A., Ostrow, A., Welihinda, A., Hayes, A., Radford, A., Mądry, A., Baker-Whitcomb, A., Beutel, A., Borzunov, A., Carney, A., Chow, A., Kirillov, Alex, Nichol, A., Paino, A., Renzin, A., Passos, A.T., Kirillov, Alexander, Christakis, A., Conneau, A., Kamali, A., Jabri, A., Moyer, A., Tam, A., Crookes, A., Tootoochian, A., Tootoonchian, A., Kumar, A., Vallone, A., Karpathy, A., Braunstein, A., Cann, A., Codispoti, A., Galu, A., Kondrich, A., Tulloch, A., Mishchenko, A., Baek, A., Jiang, A., Pelisse, A., Woodford, A., Gosalia, A., Dhar, A., Pantuliano, A., Nayak, A., Oliver, A., Zoph, B., Ghorbani, B., Leimberger, B., Rossen, B., Sokolowsky, B., Wang, B., Zweig, B., Hoover, B., Samic, B., McGrew, B., Spero, B., Giertler, B., Cheng, B., Lightcap, B., Walkin, B., Quinn, B., Guarraci, B., Hsu, B., Kellogg, B., Eastman, B., Lugaresi, C., Wainwright, C., Bassin, C., Hudson, C., Chu, C., Nelson, C., Li, C., Shern, C.J., Conger, C., Barette, C., Voss, C., Ding, C., Lu, C., Zhang, C., Beaumont, C., Hallacy, C., Koch, C., Gibson, C., Kim, C., Choi, C., McLeavey, C., Hesse, C., Fischer, C., Winter, C., Czarnecki, C., Jarvis, C., Wei, C., Koumouzelis, C., Sherburn, D., Kappler, D., Levin, D., Levy, D., Carr, D., Farhi, D., Mely, D., Robinson, D., Sasaki, D., Jin, D., Valladares, D., Tsipras, D., Li, D., Nguyen, D.P., Findlay, D., Oiwoh, E., Wong, E., Asdar, E., Proehl, E., Yang, E., Antonow, E., Kramer, E., Peterson, E., Sigler, E., Wallace, E., Brevdo, E., Mays, E., Khorasani, F., Such, F.P., Raso, F., Zhang, F., von Lohmann, F., Sulit, F., Goh, G., Oden, G., Salmon, G., Starace, G., Brockman, G., Salman, H., Bao, H., Hu, H., Wong, H., Wang, H., Schmidt, H., Whitney, H., Jun, H., Kirchner, H., Pinto, H.P. de O., Ren, H., Chang, H., Chung, H.W., Kivlichan, I., O'Connell, I., O'Connell, I., Osband, I., Silber, I., Sohl, I., Okuyucu, I., Lan, I., Kostrikov, I., Sutskever, I., Kanitscheider, I., Gulrajani, I., Coxon, J., Menick, J., Pachocki, J., Aung, J., Betker, J., Crooks, J., Lennon, J., Kiros, J., Leike, J., Park, J., Kwon, J., Phang, J., Teplitz, J., Wei, J., Wolfe, J., Chen, J., Harris, J., Varavva, J., Lee, J.G., Shieh, J., Lin, J., Yu, Jiahui, Weng, J., Tang, J., Yu, Jieqi, Jang, J., Candela, J.Q., Beutler, J., Landers, J., Parish, J., Heidecke, J., Schulman, J., Lachman, J., McKay, J., Uesato, J., Ward, J., Kim, J.W., Huizinga, J., Sitkin, J., Kraaijeveld, J., Gross, J., Kaplan, J., Snyder, J., Achiam, J., Jiao, J., Lee, J., Zhuang, J., Harriman, J., Fricke, K., Hayashi, K., Singhal, K., Shi, K., Karthik, K., Wood, K., Rimbach, K., Hsu, K., Nguyen, K., Gu-Lemberg, K., Button, K., Liu, K., Howe, K., Muthukumar, K., Luther, K., Ahmad, L., Kai, L., Itow, L., Workman, L., Pathak, L., Chen, L., Jing, L., Guy, L., Fedus, L., Zhou, L., Mamitsuka, L., Weng, L., McCallum, L., Held, L., Ouyang, L., Feuvrier, L., Zhang, L., Kondraciuk, L., Kaiser, L., Hewitt, L., Metz, L., Doshi, L., Aflak, M., Simens, M., Boyd, M., Thompson, M., Dukhan, M., Chen, Mark, Gray, M., Hudnall, M., Zhang, M., Aljubeh, M., Litwin, M., Zeng, M., Johnson, M., Shetty, M., Gupta, M., Shah, M., Yatbaz, M., Yang, M.J., Zhong, M., Glaese, M., Chen, Mianna, Janner, M., Lampe, M., Petrov, M., Wu, M., Wang, Michele, Fradin, M., Pokrass, M., Castro, M., de Castro, M.O.T., Pavlov, M., Brundage, M., Wang, Miles, Khan, M., Murati, M., Bavarian, M., Lin, M., Yesildal, M., Soto, N., Gimelshein, N., Cone, N., Staudacher, N., Summers, N., LaFontaine, N., Chowdhury, N., Ryder, N., Stathas, N., Turley, N., Tezak, N., Felix, N., Kudige, N., Keskar, N., Deutsch, N., Bundick, N., Puckett, N., Nachum, O., Okelola, O., Boiko, O., Murk, O., Jaffe, O., Watkins, O., Godement, O., Campbell-Moore, O., Chao, P., McMillan,

P., Belov, P., Su, P., Bak, P., Bakkum, P., Deng, P., Dolan, P., Hoeschele, P., Welinder, P., Tillet, Phil, Pronin, P., Tillet, Philippe, Dhariwal, P., Yuan, Q., Dias, R., Lim, R., Arora, R., Troll, R., Lin, R., Lopes, R.G., Puri, R., Miyara, R., Leike, R., Gaubert, R., Zamani, R., Wang, R., Donnelly, R., Honsby, R., Smith, R., Sahai, R., Ramchandani, R., Huet, R., Carmichael, R., Zellers, R., Chen, Roy, Chen, Ruby, Nigmatullin, R., Cheu, R., Jain, Saachi, Altman, S., Schoenholz, S., Toizer, S., Miserendino, S., Agarwal, S., Culver, S., Ethersmith, S., Gray, S., Grove, S., Metzger, S., Hermani, S., Jain, Shantanu, Zhao, S., Wu, Sherwin, Jomoto, S., Wu, Shirong, Shuaiqi, Xia, Phene, S., Papay, S., Narayanan, S., Coffey, S., Lee, S., Hall, S., Balaji, S., Broda, T., Stramer, T., Xu, T., Gogineni, T., Christianson, T., Sanders, T., Patwardhan, T., Cunninghman, T., Degry, T., Dimson, T., Raoux, T., Shadwell, T., Zheng, T., Underwood, T., Markov, T., Sherbakov, T., Rubin, T., Stasi, T., Kaftan, T., Heywood, T., Peterson, T., Walters, T., Eloundou, T., Qi, V., Moeller, V., Monaco, V., Kuo, V., Fomenko, V., Chang, W., Zheng, W., Zhou, W., Manassra, W., Sheu, W., Zaremba, W., Patil, Y., Qian, Y., Kim, Y., Cheng, Y., Zhang, Yu, He, Y., Zhang, Yuchen, Jin, Y., Dai, Y., Malkov, Y., 2024. GPT-4o System Card.

OpenAI, Achiam, J., Adler, S., Agarwal, S., Ahmad, L., Akkaya, I., Aleman, F.L., Almeida, D., Altenschmidt, J., Altman, S., Anadkat, S., Avila, R., Babuschkin, I., Balaji, S., Balcom, V., Baltescu, P., Bao, H., Bavarian, M., Belgum, J., Bello, I., Berdine, J., Bernadett-Shapiro, G., Berner, C., Bogdonoff, L., Boiko, O., Boyd, M., Brakman, A.-L., Brockman, G., Brooks, T., Brundage, M., Button, K., Cai, T., Campbell, R., Cann, A., Carey, B., Carlson, C., Carmichael, R., Chan, B., Chang, C., Chantzis, F., Chen, D., Chen, S., Chen, R., Chen, J., Chen, M., Chess, B., Cho, C., Chu, C., Chung, H.W., Cummings, D., Currier, J., Dai, Y., Decareaux, C., Degry, T., Deutsch, N., Deville, D., Dhar, A., Dohan, D., Dowling, S., Dunning, S., Ecoffet, A., Eleti, A., Eloundou, T., Farhi, D., Fedus, L., Felix, N., Fishman, S.P., Forte, J., Fulford, I., Gao, L., Georges, E., Gibson, C., Goel, V., Gogineni, T., Goh, G., Gontijo-Lopes, R., Gordon, J., Grafstein, M., Gray, S., Greene, R., Gross, J., Gu, S.S., Guo, Y., Hallacy, C., Han, J., Harris, J., He, Y., Heaton, M., Heidecke, J., Hesse, C., Hickey, A., Hickey, W., Hoeschele, P., Houghton, B., Hsu, K., Hu, S., Hu, X., Huizinga, J., Jain, Shantanu, Jain, Shawn, Jang, J., Jiang, A., Jiang, R., Jin, H., Jin, D., Jomoto, S., Jonn, B., Jun, H., Kaftan, T., Kaiser, Ł., Kamali, A., Kanitscheider, I., Keskar, N.S., Khan, T., Kilpatrick, L., Kim, J.W., Kim, C., Kim, Y., Kirchner, J.H., Kiros, J., Knight, M., Kokotajlo, D., Kondraciuk, Ł., Kondrich, A., Konstantinidis, A., Kosic, K., Krueger, G., Kuo, V., Lampe, M., Lan, I., Lee, T., Leike, J., Leung, J., Levy, D., Li, C.M., Lim, R., Lin, M., Lin, S., Litwin, M., Lopez, T., Lowe, R., Lue, P., Makanju, A., Malfacini, K., Manning, S., Markov, T., Markovski, Y., Martin, B., Mayer, K., Mayne, A., McGrew, B., McKinney, S.M., McLeavey, C., McMillan, P., McNeil, J., Medina, D., Mehta, A., Menick, J., Metz, L., Mishchenko, A., Mishkin, P., Monaco, V., Morikawa, E., Mossing, D., Mu, T., Murati, M., Murk, O., Mély, D., Nair, A., Nakano, R., Nayak, R., Neelakantan, A., Ngo, R., Noh, H., Ouyang, L., O'Keefe, C., Pachocki, J., Paino, A., Palermo, J., Pantuliano, A., Parascandolo, G., Parish, J., Parparita, E., Passos, A., Pavlov, M., Peng, A., Perelman, A., Peres, F. de A.B., Petrov, M., Pinto, H.P. de O., Michael, Pokorny, Pokrass, M., Pong, V.H., Powell, T., Power, A., Power, B., Proehl, E., Puri, R., Radford, A., Rae, J., Ramesh, A., Raymond, C., Real, F., Rimbach, K., Ross, C., Rotsted, B., Roussez, H., Ryder, N., Saltarelli, M., Sanders, T., Santurkar, S., Sastry, G., Schmidt, H., Schnurr, D., Schulman, J., Selsam, D., Sheppard, K., Sherbakov, T., Shieh, J., Shoker, S., Shyam, P., Sidor, S., Sigler, E., Simens, M., Sitkin, J., Slama, K., Sohl, I., Sokolowsky, B., Song, Y., Staudacher, N., Such, F.P., Summers, N., Sutskever, I., Tang, J., Tezak, N., Thompson, M.B., Tillet, P., Tootoonchian, A., Tseng, E., Tuggle, P., Turley, N., Tworek, J., Uribe, J.F.C., Vallone, A., Vijayvergiya, A., Voss, C., Wainwright, C., Wang, J.J., Wang, A., Wang, B., Ward, J., Wei, J., Weinmann, C., Welihinda, A., Welinder, P., Weng, J., Weng, L., Wiethoff, M., Willner, D., Winter, C., Wolrich, S., Wong, H., Workman, L., Wu, S., Wu, J., Wu, M., Xiao, K., Xu, T., Yoo, S., Yu, K., Yuan, Q., Zaremba, W., Zellers, R., Zhang, C., Zhang, M., Zhao, S., Zheng, T., Zhuang, J., Zhuk,

W., Zoph, B., 2023. GPT-4 Technical Report.

Periti, F., Cassotti, P., Dubossarsky, H., Tahmasebi, N., 2024. Analyzing Semantic Change through Lexical Replacements, in: Proceedings of the 62nd Annual Meeting of the Association for Computational Linguistics (Volume 1: Long Papers). pp. 4495–4510.

Periti, F., Montanelli, S., 2024. Lexical Semantic Change through Large Language Models: a Survey. ACM Comput. Surv. 56. https://doi.org/10.1145/3672393

Schick, T., Schütze, H., 2020. It's Not Just Size That Matters: Small Language Models Are Also Few-Shot Learners. NAACL-HLT 2021 - 2021 Conf. North Am. Chapter Assoc. Comput. Linguist. Hum. Lang. Technol. Proc. Conf. 2339–2352. https://doi.org/10.18653/v1/2021.naacl-main.185

Shen, S., Logeswaran, L., Lee, M., Lee, H., Poria, S., Mihalcea, R., 2024. Understanding the Capabilities and Limitations of Large Language Models for Cultural Commonsense, in: Proceedings of the 2024 Conference of the North American Chapter of the Association for Computational Linguistics: Human Language Technologies, NAACL 2024. Association for Computational Linguistics (ACL), pp. 5668–5680. https://doi.org/10.18653/v1/2024.naacl-long.316

Spataru, A., Hambro, E., Voita, E., Cancedda, N., 2024. Know When To Stop: A Study of Semantic Drift in Text Generation, in: Proceedings of the 2024 Conference of the North American Chapter of the Association for Computational Linguistics: Human Language Technologies, NAACL 2024. Association for Computational Linguistics (ACL), pp. 3656–3671. https://doi.org/10.18653/v1/2024.naacl-long.202

Tao, Y., Viberg, O., Baker, R.S., Kizilcec, R.F., 2023. Cultural Bias and Cultural Alignment of Large Language Models. Acad. Tao, O Viberg, RS Bak. RF KizilcecPNAS nexus, 2024•academic.oup.com.

Tripodi, R., Warglien, M., Levis Sullam, S., Paci, D., 2019. Tracing Antisemitic Language Through Diachronic Embedding Projections: France 1789-1914. Association for Computational Linguistics (ACL), pp. 115–125. https://doi.org/10.18653/v1/w19-4715

Vijayarani, J., Geetha, T. V., 2020. Knowledge-enhanced temporal word embedding for diachronic semantic change estimation. Soft Comput. 24, 12901–12918. https://doi.org/10.1007/s00500-020-04714-0

Wegmann, A., Lemmerich, F., Strohmaier, M., 2020. Detecting Different Forms of Semantic Shift in Word Embeddings via Paradigmatic and Syntagmatic Association Changes, in: Lecture Notes in Computer Science (Including Subseries Lecture Notes in Artificial Intelligence and Lecture Notes in Bioinformatics). Springer Science and Business Media Deutschland GmbH, pp. 619–635. https://doi.org/10.1007/978-3-030-62419-4_35

Wevers, M., Koolen, M., 2020. Digital begriffsgeschichte: Tracing semantic change using word embeddings. Hist. Methods 53, 226–243. https://doi.org/10.1080/01615440.2020.1760157

Yang, K., Liu, J., Wu, J., Yang, C., Fung, Y.R., Li, S., Huang, Z., Cao, X., Wang, X., Wang, Y., Ji, H., Zhai, C., 2024. If LLM Is the Wizard, Then Code Is the Wand: A Survey on How Code Empowers Large Language Models to Serve as Intelligent Agents.

Zhao, W.X., Zhou, K., Li, J., Tang, T., Wang, X., Hou, Y., Min, Y., Zhang, B., Zhang, J., Dong, Z., Du, Y., Yang, C., Chen, Y., Chen, Z., Jiang, J., Ren, R., Li, Y., Tang, X., Liu, Z., Liu, P., Nie, J.-Y., Wen, J.-R., 2023. A Survey of Large Language Models.

한국컴퓨터종합학술대회 논문집.